\documentclass[runningheads]{meta}

\usepackage{epsfig}
\usepackage{esvect}

\usepackage{algorithmic}
\usepackage[ruled,vlined,linesnumbered]{algorithm2e}

{}

{}
{}
{}

\newcommand{\trp}[2]{\paths{#1}{1}{#2}}{}
\newcommand{\paths}[3]{p_{#1,\vv{q_{#2} q_{#3}}}}{}

{}

\newcommand{\delt}[3]{\delta_{#1,\vv{q_{#2} q_{#3}}}}{}
\newcommand{\Sp}{S^{+}}{}
\newcommand{\Sm}{S^{-}}{}
\newcommand{\ders}[3]{D_{{#1},\vv{q_{#2} q_{#3}}}}{}
{}

{}

\usepackage{xcolor}
\usepackage{colortbl}

\begin{document}

\pagestyle{headings}

\mainmatter

\title{GA and ILS for optimizing the size of NFA models}

\titlerunning{GA and ILS for optimizing the size of NFA models}

\author{Frédéric Lardeux\inst{1} 
\and
Eric Monfroy\inst{1} 
}

\authorrunning{Lardeux and Monfroy}

\institute{Univ Angers, LERIA, SFR MATHSTIC, F-49000 Angers, France\\
\email{firstname.lastname@univ-angers.fr}}

\maketitle

\begin{abstract}
Grammatical inference consists in learning a formal grammar (as a set of rewrite rules or a finite state machine). We are concerned with learning Nondeterministic Finite Automata (NFA) of a given size from samples of positive and negative words. NFA can naturally be modeled in SAT. The standard model~\cite{WieczorekBook} being enormous, we also try a model based on prefixes~\cite{DBLP:journals/fuin/JastrzabCW21} which generates smaller instances. We also propose a new model based on suffixes and a hybrid model based on prefixes and suffixes. We then focus on optimizing the size of generated SAT instances issued from the hybrid models. 
%
We present two techniques to optimize this combination, one based on Iterated Local Search (ILS), the second one based on Genetic Algorithm (GA). 
Optimizing the combination significantly reduces the SAT instances and their solving time, but at the cost of longer generation time. 
We, therefore, study the balance between generation time and solving time thanks to some experimental comparisons, and we analyze our various model improvements.

\end{abstract}
\keywords{Constraint problem modeling, Grammar inference, SAT, model reformulation, NFA inference.}

\section{Introduction}
\label{sec:Introduction}

Grammatical inference~\cite{ColinBook} (or grammar induction) is concerned with the study of algorithms for learning automata and grammars from some observations. The goal is thus to construct a representation that accounts for the characteristics of the observed objects. This research area plays a significant role in numerous applications, such as compiler design, bioinformatics,  speech recognition, 
pattern recognition, machine learning, and others. 

In this article, we focus on learning a finite automaton from samples of words $S=\Sp \cup \Sm$, such that $\Sp$ is a set of positive words that must be accepted by the automaton, and $\Sm$ is a set of negative words to be rejected by the automaton. Due to their determinism, deterministic finite automata (DFA) are generally faster than non deterministic automata (NFA). However, NFA are significantly smaller than DFA in terms of the number of states. Moreover, the space complexity of the SAT models representing the problem is generally due to the number of states. Thus, we focus here on NFA inference. An NFA is represented by a 5-tuple $(Q , \Sigma , \Delta, q_1, F )$ where $Q$ is a finite set of states, the vocabulary $\Sigma$ is a finite set of symbols, the transition function $\Delta  : Q \times \Sigma \rightarrow {\mathcal P}(Q)$ associates a set of states to a given state and a given symbol, $q_1 \in Q$ is the initial state, and $F \subseteq Q$ is the set of final states.

The problem of inferring NFA has been undertaken with various approaches (see, e.g., \cite{WieczorekBook}).
%
Among them,
we can cite ad-hoc algorithms such as \textit{DeLeTe2}~\cite{delete2} that is based on state merging methods, or the technique of~\cite{DBLP:conf/wia/PargaGR06} that returns a collection of NFA. Some approaches use metaheuristics for computing NFA, such as hill-climbing~\cite{tomita82} or genetic algorithm~\cite{DBLP:conf/icgi/Dupont94}.
%
%
%

A convenient and declarative way of representing combinatorial problems is to model them as a Constraint Satisfaction Problem (CSP~\cite{Rossi2006}) (see, e.g.,~\cite{WieczorekBook} for an INLP model for inferring NFA, or~\cite{ola2021} for a SAT (the propositional satisfiability problem~\cite{Garey1979}) model of the same problem). 
Parallel solvers have also been used for minimizing the inferred NFA size~\cite{jastrzab2017,DBLP:journals/fuin/JastrzabCW21}. 

Orthogonally to the approaches cited above, we do not seek to improve a solver, but to generate a model of the problem that is easier to solve with a standard SAT solver. Our approach is similar to DFA inference with graph coloring~\cite{HeuleMarijn2013Smsu}, or NFA inference with complex data structures~\cite{ola2021}. 
Modeling thus consists in translating a problem into a CSP made of decision variables and constraints over these variables. 
As a reference for comparisons, we start with the basic SAT model of~\cite{ola2021}. 
%
The model, together with a sample of positive and negative words, lead to a SAT instance to be solved by a classic SAT solver that we use as a black box. However, SAT instances are gigantic, e.g., our base model space complexity is in the order of ${\mathcal{O}}(k^{|\omega_{+}|})$ variables, and in $\mathcal{O}(|\omega_{+}|.k^{|\omega_{+}|})$ clauses, where $k$ is the number of states of the NFA, and $\omega_{+}$ is the size of the longest positive word of the sample.
The second model, $PM$, is based on intermediate variables for each prefix~\cite{DBLP:journals/fuin/JastrzabCW21} which enables to compute only once parts of paths that are shared by several words. We propose a third model, $SP$, based on intermediate variables for suffixes. Although the two models could seem similar, their order of size is totally different. Indeed, $PM$ is in $\mathcal{O}(k^2)$ while $SM$ is in $\mathcal{O}(k^3)$.
We then propose hybrid models consisting in splitting words into a prefix and a suffix. Modeling the beginning of the word is made with $PM$ while the suffix is modeled by $SM$. The challenge is then to determine where to split words to optimize the size of the generated SAT instances. To this end, we propose two approaches, one based on iterated local search (ILS), the second one on genetic algorithm (GA). Both permit to generate smaller SAT instances, much smaller than with the $DM$ model and even the $PM$ model. However, with GA, the generation time is too long and erases the gain in solving with the Glucose SAT solver~\cite{glucose}. But the hybrid instances optimized with the ILS are smaller, and the generation time added to the solving time is faster than with $PM$.
Compared to~\cite{ola2021}, which is the closest work on NFA inferring, we always obtain significantly smaller instances and solving time.

This paper is organized as follows. In Section~\ref{sec:SATmodels} we present the direct model, the prefix model, and we propose the suffix model. We then combine suffix and prefix model to propose the new hybrid models (Section~\ref{subsec:HM}). Hybrid models are optimized with iterated local search (Sub-section~\ref{subsec:ILSHM}), and with genetic algorithm in Sub-section~\ref{subsec:GAHM}. We then compare experimentally our models  in Section~\ref{sec:Expe} before concluding in Section~\ref{sec:Conclusion}.

\section{SAT Models}
\label{sec:SATmodels}

Given
an alphabet $\Sigma=\{s_1,\ldots,s_n\}$ of $n$ symbols,
    a training sample $S=\Sp \cup \Sm$, where $\Sp$ (respectively $\Sm$) is a set of \textit{positive words} (respectively \textit{negative words}) from $\Sigma^{*}$,
    and an integer $k$,   
\textbf{the NFA inference problem} consists in building a NFA with $k$ states which validates words of $\Sp$, and rejects words of $\Sm$. 
Note that the satisfaction problem we consider in this paper can be extended to an optimization problem minimizing $k$~\cite{DBLP:journals/fuin/JastrzabCW21}. 

Let us introduce some notations. 
Let $A=(Q,\Sigma, q_1, F)$ be a NFA with:
 $Q=\{q_1,\ldots,q_k\}$ a set of $k$ states,
    $\Sigma$ a finite alphabet,
    $q_1$ the initial state,
    and $F$ the set of final states.
The empty word is noted $\lambda$. We denote by $K$ the set of integers $\{1, \ldots,k\}$.
%
%

We consider the following variables:
\begin{itemize}
    \item $k$ the size of the NFA we want to learn,
    \item a set of $k$ Boolean variables $F=\{f_1, \ldots, f_k\}$  determining whether states $q_1$ to $q_k$ are final or not,
    \item and $\Delta=\{\delt{s}{i}{j}| s \in \Sigma \textrm{~and~} i,j \in K\}$ a set of $n.k^2$ Boolean variables defining the existence or not of the transition from state $q_i$ to state $q_j$ with the symbol $s$, for each $q_i$, $q_j$, and $s$.
\end{itemize}    

The path $i_1, i_2, \ldots, i_{n+1}$ for $w=w_1 \ldots  w_n$ exists if and only if $d=\delt{w_1}{i_1}{i_2} \wedge  \ldots \wedge \delt{w_n}{i_{n}}{i_{n+1}}$ is true. We say that the conjunction $d$ is a c\_path, and $\ders{w}{i}{j}$ is the set of all c\_paths for the word $w$ between states $q_i$ and $q_j$.

%
\subsection{Direct Model}
\label{subsec:DM}

This simple model has been presented in~\cite{ola2021}. It is based on 3 sets of equations:
\begin{enumerate}
    \item If the empty word is in $\Sp$ or $\Sm$, we can fix whether the first state is final or not:
    \begin{eqnarray}
    \textrm{if } \lambda \in \Sp, ~~~~~~f_1 \label{lambda1} \\
    \textrm{if } \lambda \in \Sm, ~~~~    \neg f_1 \label{lambda2}
    \end{eqnarray}    
    
    \item For each word $w \in \Sp$, there is at least a path from $q_1$ to a final state $q_j$:
    \begin{eqnarray}
     \bigvee_{j \in K} \bigvee_{~d \in \ders{w}{1}{j}} \big( d \wedge f_j \big) \label{m1}
    \end{eqnarray}
    
    With the Tseitin transformations~\cite{Tseitin1983}, we create one auxiliary variable for each combination of a word $w$, a state $j \in K$, and a c\_path $d \in \ders{w}{1}{j}$: 
    $
    aux_{w,j,d} \leftrightarrow d \wedge f_j 
    $.
    Hence, we obtain a formula in CNF for each $w$:
    \begin{eqnarray}
    \bigwedge_{j \in K} 
    \bigwedge_{~d \in \ders{w}{1}{j}}
    \left[
    (\neg aux_{w,j,d} \vee (d \wedge f_j)) 
    \right]  \label{aux1Mk} \\
    %
    \bigwedge_{j \in K} 
    \bigwedge_{~d \in \ders{w}{1}{j}}
    (aux_{w,j,d} \vee \neg d \vee \neg f_j)   \label{aux2Mk} \\
    %
    \bigvee_{j \in K} 
    \bigvee_{~d \in \ders{w}{1}{j}}
        aux_{w,j,d} \label{aux3Mk}
    \end{eqnarray}
    

    \item For each $w \in \Sm$ and each $q_j$, either there is no path  state $q_1$ to $q_j$, or $q_j$ is not final:
    \begin{eqnarray}
    \neg \left[
     \bigvee_{j \in K} \bigvee_{~d \in \ders{w}{1}{j}} \big(d  \wedge f_j \big) 
    \right] 
    \label{negM}
    \end{eqnarray}
    \end{enumerate}

    Thus, the direct constraint model $DM_k$ for building a NFA of size $k$ is: 
    \[DM_k=   \bigwedge_{w \in \Sp} 
    \Big((\ref{aux1Mk}) \wedge 
    (\ref{aux2Mk}) \wedge 
    (\ref{aux3Mk}) \Big) 
    \wedge 
    \bigwedge_{w \in \Sm} (\ref{negM})\]
    and is possibly completed by $(\ref{lambda1})$ or $(\ref{lambda2})$ if $\lambda \in \Sp$ or $\lambda \in \Sm$.

\paragraph{\textbf{Size of the models} (see~\cite{ola2021} for details)}

Consider $\omega_{+}$ and $\omega_{-}$, the longest word of $\Sp$ and $\Sm$ respectively. Table~\ref{clDM} presents the number of clauses (Column 1) and their arities (Column 2), which are an upper bound of a given constraint group (last column) for the model $SM_{k}$. Table~\ref{varDM} presents the upper bound of the number of Boolean variables that are required and why the are required.
We can see on Tables~\ref{clDM} and~\ref{varDM} that the space complexity of the $DM_{k}$ is huge (${\mathcal{O}}(|\Sp|.k.^{|\omega_{+}|})$ variables, and $\mathcal{O}(|\Sp|.(|\omega_{+}|+1).k^{|\omega_{+}|})$ clauses) and with large clauses (up to arity of $|\omega_{+}|+2$), and that only small instances for a small number of states will be tractable. It is thus obvious that it is important to improve the model $DM_k$.

%
%

\begin{table}[t]
        \begin{minipage}[t]{0.45\textwidth}
            \centering
            $\begin{array}{|l|r|c|}
            \hline
            \textrm{number of cl.} & \textrm{arity} & \textrm{Constraints} \\ \hline
            |\Sp|.(|\omega_{+}|+1).k^{|\omega_{+}|}   &  2  & (\ref{aux1Mk})\\
            |\Sp|.k^{|\omega_{+}|} & |\omega_{+}|+2 & (\ref{aux1Mk})\\
            |\Sp| & k^{|\omega_{+}|} & (\ref{aux1Mk})\\

            |\Sm|.k^{|\omega_{-}|} & |\omega_{-}|+1 & (\ref{negM})\\   

            \hline
            \end{array}$
            
            \caption{Clauses for $DM_k$\label{clDM}}
        \end{minipage}
            \begin{minipage}[t]{0.45\textwidth}
            \centering
            $\begin{array}{|l|l|}
            \hline
            \textrm{number of var} & \textrm{reason} \\ \hline
            k & \textrm{final states }F \\
            n.k^2 & \textrm{transitions } \delta \\
            |\Sp|.k.^{|\omega_{+}|} & \textrm{Constraints (\ref{m1})}  \\ \hline
            \end{array}$
    
            \caption{Variables for $DM_k$\label{varDM}}
        \end{minipage}
\end{table}

\subsection{Prefix Model~\cite{DBLP:journals/fuin/JastrzabCW21}}
\label{subsec:PM}

Let $Pref(w)$ be the set of all the non-empty prefixes of the word $w$ and, by extension, $Pref(W)=\cup_{w \in W}Pref(w)$ the set of prefixes of the words of the set $W$.
%
For each $w \in Pref(S)$, we add a Boolean variable $\trp{w}{i}$ which determines whether there is or not a c\_path for $w$ from state $q_1$ to $q_i$. Note that these variables can be seen as labels of the Prefix Tree Acceptor (PTA) for $S$~\cite{ColinBook}.
%
%
%
The problem can be modeled with the following constraints:
\begin{enumerate}
    
    \item For all prefix $w=a$ with $w\in Pref(S)$, and $a \in \Sigma$, there is a c\_path of size 1 for $w$:
    \begin{eqnarray}
    \bigvee_{i \in K} \delt{a}{1}{i} \leftrightarrow \trp{a}{i} \label{pref0}
    \end{eqnarray}
    
    With the Tseitin transformations, we can derive a CNF formula. It is also possible to directly encode $\delt{a}{1}{i}$ and $\trp{a}{i}$ as the same variable. Thus, no clause is required.
    
    \item For all words $w \in \Sp-\{\lambda\}$:
    \begin{eqnarray}
    \bigvee_{i \in K} \trp{w}{i} \wedge f_i \label{pref1}
    \end{eqnarray}
    
    With the Tseitin transformations~\cite{Tseitin1983}, we create one auxiliary variable for each combination of 
    $\trp{w}{i}$ and the status (final or not) of the state $q_i$: 
    $
    aux_{w,i} \leftrightarrow \trp{w}{i} \wedge f_i \nonumber
    $.
    Hence, for each $w$, we obtain a formula in CNF:
    \begin{eqnarray}
    \bigwedge_{i \in K} 
    ((\neg aux_{w,i} \vee \trp{w}{i}) \wedge (\neg aux_{w,i} \vee f_i))
      \label{aux1pref} \\
    \bigwedge_{i \in K} 
    (aux_{w,i} \vee \neg \trp{w}{i} \vee \neg f_i)
      \label{aux2pref} \\
    %
    \bigvee_{i \in K} 
        aux_{w,i} \label{aux3pref}
    \end{eqnarray}


    \item For all words $w \in \Sm-\{\lambda\}$, we obtain the following CNF constraint:
    \begin{eqnarray}
    \bigwedge_{i \in K} (\neg \trp{w}{i} \vee \neg f_i) \label{negpref2}
    \end{eqnarray}
    


    \item For all prefix $w=va$, $w\in Pref(S)$, 
    $v \in Pref(S)$
    and $a \in \Sigma$:
    \begin{eqnarray}
    \bigwedge_{i \in K} (\trp{w}{i} \leftrightarrow (\bigvee_{j \in K} \trp{v}{j} \wedge  \delt{a}{j}{i})) \label{prefrec}
    \end{eqnarray}
    
    Applying the Tseitin transformations, we create one auxiliary variable for each combination of existence of a c\_path from $q_1$ to $q_i$ ($\trp{v}{j}$) and the transition $\delt{a}{j}{i}$: 
    $
    aux_{v,a,j,i} \leftrightarrow \trp{v}{j} \wedge \delt{a}{j}{i} \nonumber
    $.
    Then, (\ref{prefrec}) becomes:
    \begin{eqnarray}
    \bigwedge_{i \in K} (\trp{w}{i} \leftrightarrow (\bigvee_{j \in K} aux_{v,a,j,i}))\nonumber
    \end{eqnarray}

    For each $w \in Pref(S)$, we obtain constraints in CNF:
    \begin{eqnarray}
    \bigwedge_{(i,j) \in K^2} \ (\neg aux_{v,a,j,i} \vee \trp{w}{i}) \label{auxprefrec1}\\
    \bigwedge_{(i,j) \in K^2} (\neg aux_{v,a,j,i} \vee \delt{a}{j}{i}) \label{auxprefrec2}\\
    \bigwedge_{(i,j) \in K^2} (aux_{v,a,j,i} \vee \neg \trp{w}{i} \vee \neg \delt{a}{j}{i}) \label{auxprefrec3}\\
    \bigwedge_{i \in K} (\neg \trp{w}{i} \vee (\bigvee_{j \in K} aux_{v,a,j,i})) \label{auxprefrec4}\\
    \bigwedge_{(i,j) \in K^2} (\trp{w}{i} \vee \neg aux_{v,a,j,i})) \label{auxprefrec5}
    \end{eqnarray}
  

    \end{enumerate}

    Thus, the constraint prefix model $PM_{k}$ for building a NFA of size $k$ is: 
    \[PM_{k}=   \bigwedge_{w \in \Sp} 
    \Big((\ref{aux1pref}) \wedge 
    \ldots  \wedge
    (\ref{aux3pref}) \Big) 
     \wedge
    \bigwedge_{w \in \Sm} 
    (\ref{negpref2}) 
     \wedge
    \bigwedge_{w \in Pref(S)} 
    (\ref{auxprefrec1}) \wedge \ldots \wedge (\ref{auxprefrec5})
    \]
    and is possibly completed by $(\ref{lambda1})$ or $(\ref{lambda2})$ if $\lambda \in \Sp$ or $\lambda \in \Sm$.

\paragraph{\textbf{Size of the models}}

Consider $\omega_{+}$, the longest word of $\Sp$, $\omega_{-}$, the longest word of $\Sm$, $\sigma=\Sigma_{w\in S}|w|$, and $\pi$, the number of prefix obtained by $Pref(S)$ with a size larger than 1 ($\pi=|\{x|x\in Pref(S), |x|>1\}|$), then:
$$ max(|\omega_+|,|\omega_-|) \leq \pi \leq \sigma \leq |\Sp|.|\omega_+|+ |\Sm|.|\omega_-|$$
The space complexity of the $PM_{k}$ model is thus in 
${\mathcal{O}}(\sigma.k^2)$ variables, and in ${\mathcal{O}}(\sigma.k^2)$ binary and ternary clauses, and ${\mathcal{O}}(\sigma.k)$ ($k+1$)-ary clauses.

    

\begin{table}[t]
        \begin{minipage}[t]{0.45\textwidth}
            \centering
            $\begin{array}{|l|r|c|}
            \hline
            \textrm{number of cl.} & \textrm{arity} & \textrm{Constraints} \\ \hline
            2.k.|\Sp|   &  2  & (\ref{aux1pref})\\
            k.|\Sp| & 3 & (\ref{aux2pref})\\
            k & k+1 & (\ref{aux3pref})\\

            k.|\Sm| & 2 & (\ref{negpref2})\\   

            \pi.k^2 & 2  &   (\ref{auxprefrec1})\\
            \pi.k^2 & 2  &   (\ref{auxprefrec2})\\
            \pi.k^2 & 3  &   (\ref{auxprefrec3})\\
            \pi.k & k+1  &   (\ref{auxprefrec4})\\
            \pi.k^2 & 2  &   (\ref{auxprefrec5})\\
            \hline
            \end{array}$
            
            \caption{Clauses for $PM_k$\label{clPM}}
        \end{minipage}
            \begin{minipage}[t]{0.45\textwidth}
            \centering
            $\begin{array}{|l|l|}
            \hline
            \textrm{number of var} & \textrm{reason} \\ \hline
            k & \textrm{final states }F \\
            n.k^2 & \textrm{transitions } \delta \\
            |\Sp|.k & \textrm{Constraints (\ref{pref1})} \\
            \pi.k^2 & \textrm{Constraints (\ref{prefrec})} \\ \hline
            \end{array}$
    
            \caption{Variables for $PM_k$\label{varPM}}
        \end{minipage}
\end{table}    

\subsection{Suffix Model}
\label{subsec:RM}

We now propose a suffix model ($SM_k$), based on $Suf(S)$, the set of all the non-empty suffixes of all the words in $S$. The main difference is that the construction starts from every state and terminates in state $q_1$. 
For each $w \in Suf(S)$, we add a Boolean variable $\paths{w}{i}{j}$ which determines whether there is or not a c\_path for $w$ from state $q_i$ to $q_j$.
%
%
To model the problem, Constraints~(\ref{aux1pref}), (\ref{aux2pref}), (\ref{aux3pref}), and (\ref{negpref2}) remain unchanged and creation of the corresponding auxiliary variables $aux_{w,i}$ as well.

    \medskip

    For each suffix $w=a$ with $w\in Suf(S)$, and $a \in \Sigma$, there is a c\_path of size 1 for $w$:
    \begin{eqnarray}
    \bigvee_{(i,j) \in K^2} \delt{a}{i}{j} \leftrightarrow \paths{a}{i}{j} \label{suf0}
    \end{eqnarray}
    We can directly encode $\delt{a}{i}{j}$ and $\paths{a}{i}{j}$ as the same variable. Thus, no clause is required.

    \medskip
    
    For all suffix $w=av$, $w\in Suf(S)$, $v \in Suf(S)$ and $a \in \Sigma$:
    \begin{eqnarray}
    \bigwedge_{(i,j) \in K^2} (\paths{w}{i}{j} \leftrightarrow 
    (\bigvee_{k \in K} \delt{a}{i}{k}  \wedge  \paths{v}{k}{j}   )) \label{sufrec}
    \end{eqnarray}

    We create one auxiliary variable for each combination of existence of a c\_path from $q_k$ to $q_j$ ($\paths{v}{k}{j}$) and the transition $\delt{a}{i}{k}$: 
    $
    aux_{v,a,i,k,j} \leftrightarrow \delt{a}{i}{k}  \wedge  \paths{v}{k}{j}  \nonumber
    $
    

    For each $w=av$, we obtain the following constraints (CNF formulas):
    \begin{eqnarray}
    \bigwedge_{(i,j,k) \in K^3} (\neg aux_{v,a,i,k,j} \vee \paths{w}{k}{j}) \label{auxsufrec1}\\
    \bigwedge_{(i,j,k) \in K^3} (\neg aux_{v,a,i,k,j} \vee \delt{a}{i}{k}) \label{auxsufrec2}\\
    \bigwedge_{(i,j,k) \in K^3} (aux_{v,a,i,k,j} \vee \neg \paths{w}{k}{j} \vee \neg \delt{a}{i}{k}) \label{auxsufrec3}\\
    \bigwedge_{(i,j) \in K^2} (\neg \paths{w}{i}{j} \vee (\bigvee_{k \in K} aux_{v,a,i,k,j})) \label{auxsufrec4}\\
    \bigwedge_{(i,j,k) \in K^3} (\paths{w}{i}{j} \vee \neg aux_{v,a,i,k,j})) \label{auxsufrec5}
    \end{eqnarray}
  
   \medskip
   
   Note that some clauses are not worth being generated. Indeed, it is useless to generate paths starting in states different from the initial state $q_1$, except when the $w$ is in $S$, and $w$ is also the suffix of another word from $S$. Removing these constraints does not change the complexity of the model. This can easily be done at generation time, or we can leave it to the solver, which will detect it and remove the useless constraints.

    Thus, the constraint prefix model $PM_{k}$ for building a NFA of size $k$ is: 
    \[SM_{k}=   \bigwedge_{w \in \Sp} 
    \Big((\ref{aux1pref}) \wedge 
    \ldots  \wedge
    (\ref{aux3pref}) \Big) 
     \wedge
    \bigwedge_{w \in \Sm} 
    (\ref{negpref2}) 
     \wedge
    \bigwedge_{w \in Pref(S) \setminus S} 
    (\ref{auxsufrec1}) \wedge \ldots \wedge (\ref{auxsufrec5}) 
    \]
    and is possibly completed by $(\ref{lambda1})$ or $(\ref{lambda2})$ if $\lambda \in \Sp$ or $\lambda \in \Sm$.

\paragraph{\textbf{Size of the models}}

Consider $\omega_{+}$, $\omega_{-}$, $\sigma$, and $\pi$ as defined in the prefix model. Table~\ref{clSM} presents the number of clauses (first column) and their arities (Column 2) which are an upper bound of a given constraint group (last column) for the model $SM_{k}$. Table~\ref{varSM} presents the upper bound of the number of Boolean variables that are required, and the reason of their requirements. To simplify, the space complexity of $SM_k$ is thus in ${\mathcal{O}}(\sigma.k^3)$ variables,  and in ${\mathcal{O}}(\sigma.k^3)$ binary and ternary clauses, and ${\mathcal{O}}(\sigma.k^2)$ ($k+1$)-ary clauses.
\begin{table}[t]
        \begin{minipage}[t]{0.45\textwidth}
            \centering
            $\begin{array}{|l|r|c|}
            \hline
            \textrm{number of cl.} & \textrm{arity} & \textrm{Constraints} \\ \hline
            2.k.|\Sp|   &  2  & (\ref{aux1pref})\\
            k.|\Sp| & 3 & (\ref{aux2pref})\\
            k & k+1 & (\ref{aux3pref})\\

            k.|\Sm| & 2 & (\ref{negpref2})\\   

            \pi.k^3 & 2  &   (\ref{auxsufrec1})\\
            \pi.k^3 & 2  &   (\ref{auxsufrec2})\\
            \pi.k^3 & 3  &   (\ref{auxsufrec3})\\
            \pi.k^2 & k+1  &   (\ref{auxsufrec4})\\
            \pi.k^3 & 2  &   (\ref{auxsufrec5})\\
            \hline
            \end{array}$
            
            \caption{Clauses for $SM_k$\label{clSM}}
        \end{minipage}
            \begin{minipage}[t]{0.45\textwidth}
            \centering
            $\begin{array}{|l|l|}
            \hline
            \textrm{number of var} & \textrm{reason} \\ \hline
            k & \textrm{final states }F \\
            n.k^2 & \textrm{transitions } \delta \\
            |\Sp|.k & \textrm{Constraints (\ref{pref1})} \\
            \pi.k^3 & \textrm{Constraints (\ref{sufrec})} \\ \hline
            \end{array}$
    
            \caption{Variables for $SM_k$\label{varSM}}
        \end{minipage}
\end{table}


\section{Hybrid Models}
\label{subsec:HM}
We now propose a family of models based on both the notion of prefix and the notion of suffix. The idea is, in fact, to take advantage of the construction of a prefix $p$ and a suffix $s$ of a word $w$ such that $w=p.s$ to pool both prefixes and suffixes. The goal is to reduce the size of generated SAT instances.
The process is the following:
\begin{enumerate}
    \item For each word $w_i$ of $S$, we split $w_i$ into $p_i$ and $s_i$ such that $w=p_i.s_i$. We thus obtain two sets, $S_p=\{p_i ~|~ \exists i, w_i \in S  \textrm{ and } w_i=p_i.s_i\}$ and $S_s=\{s_i ~|~ \exists i, w_i \in S \textrm{ and } w_i=p_i.s_i\}$.
    
    \item We then consider $S_p$ as a sample, i.e., a set of words. For each $w$ of $S_p$, we generate Constraints (\ref{auxprefrec1}) to (\ref{auxprefrec5}).
    
    \item We consider $S_s$ in turn to generate Constraints (\ref{auxsufrec1}) to (\ref{auxsufrec5}) for each $w \in S_s$.
    
    \item Then, for each $w_i=p_i.s_i$, clauses corresponding to $p_i$ must be linked to clauses of $s_i$.
    \begin{itemize}
        \item if $w_i=p_i.s_i \in \Sm$, the constraints are similar to the ones of~(\ref{negpref2}) including the connection of $p_i$ and $s_i$:
    
            \begin{eqnarray}
            \bigwedge_{(j,k) \in K^2} (\neg \paths{p_i}{1}{j} \vee \neg \paths{s_i}{j}{k} \vee \neg f_i) \label{hybneg}
            \end{eqnarray}
    
        \item if $w_i=p_i.s_i \in \Sp$, the constraints are similar to~( \ref{pref1}):
        \begin{eqnarray}
            \bigvee_{(j,k) \in K^2} \paths{p_i}{1}{j} \wedge \paths{s_i}{j}{k} \wedge f_k \label{hybpref1}
        \end{eqnarray}
        We transform (\ref{hybpref1}) using auxiliary variables $aux_{w_i,j,k} \leftrightarrow \paths{w}{1}{j} \wedge \paths{w}{j}{k} \wedge f_i$ to obtain the following CNF constraints:
        
    \begin{eqnarray}
    \bigwedge_{(j,k) \in K^2} 
    ((\neg aux_{w_i,j,k} \vee \paths{w}{1}{j}) \wedge (\neg aux_{w_i,j,k} \vee \paths{w}{j}{k}) \wedge (\neg aux_{w_i,j,k} \vee f_k))
      \label{hybaux1pref} \\
    \bigwedge_{(j,k) \in K^2}
    (aux_{w_i,j,k} \vee \neg \paths{w}{1}{j} \vee \paths{w}{j}{k} \vee \neg f_k)
      \label{hybaux2pref} \\
    %
    \bigvee_{(j,k) \in K^2} 
        aux_{w_i,j,k} \label{hybaux3pref}
    \end{eqnarray}
    \end{itemize}
    
\end{enumerate}

    Thus, the hybrid model $HM_{k}$ for building a NFA of size $k$ is: 
    \[HM_{k}=   \bigwedge_{w \in \Sp} 
    \Big((\ref{hybaux1pref}) \wedge 
    \ldots  \wedge
    (\ref{hybaux3pref}) \Big) 
     \wedge
    \bigwedge_{w \in \Sm} 
    (\ref{hybneg}) 
     \wedge
    \bigwedge_{p_i \in Pref(S_p)} 
    (\ref{auxsufrec1}) \wedge \ldots \wedge (\ref{auxsufrec5}) 
        \bigwedge_{s_i \in Suf(S_s)} 
    (\ref{auxprefrec1}) \wedge \ldots \wedge (\ref{auxprefrec5}) 
    \]
    and it is possibly completed by $(\ref{lambda1})$ or $(\ref{lambda2})$ if $\lambda \in \Sp$ or $\lambda \in \Sm$.

We do not detail it here, but in the worst case, the complexity of the model is the same as $SM_k$.
It is obvious that the split of each word into a prefix and a suffix will determine the size of the instance. The next sub-sections are dedicated to the computation of this separation $w_i=p_i.s_i$ to minimize the size of the generated hybrid instances with the $HM_k$ model.


\subsection{Search Space and Evaluation Function For Metaheuristics}
\label{subsec:SSandEF}
The search space $\cal X$ of this problem corresponds to all the hybrid models: for each word $w$ of $S$, we have to determine a $n$ such that $w=p.s$ with $|p|=n$ and $|s|=|w|-n$. The size of the search space is thus: $|{\cal X}| = \Pi_{w\in S}|w|+1$.

Even though we are aware that smaller instances are not necessarily easier to solve, we choose to define the first evaluation function as the number of generated SAT variables. However, this number cannot be computed a priori: first, the instance has to be generated, before counting the variables.
This function being too costly, we propose an alternative evaluation function for approximating the number of variables. This fitness function is 
based on the number of prefixes in $Pref(S_p)$ and suffixes in $Suf(S_s)$. 
Since the complexity of $SM_k$ is in ${\cal O}(k^3)$ whereas the complexity of $PM_k$ is in ${\cal O}(k^2)$, suffixes are penalized by a coefficient corresponding to the number of states. 

$$fitness(S_p,S_s)=|Pref(S_p)|+k.|Suf(S_s)|$$
Empirically, we observe that the results of this $fitness$ function are proportional to the actual number of generated SAT variables. This approximation of the number of variables will thus be the fitness function in our ILS and GA algorithms.

\subsection{Iterated Local Search Hybrid Model $HM\_ILS_k$}
\label{subsec:ILSHM}

We propose an Iterated Local Search (ILS) \cite{Stutzle2018} for optimizing our hybrid model. 
Classically, a best improvement or a first improvement neighborhood is used in ILS to select the next move. In our case, a first improvement provides very poor results. Moreover, it is clearly impossible to evaluate all the neighbors at each step due to the computing cost. We thus decide to randomly choose a word in $S$ with a roulette wheel selection based on the word weights. Each word $w$ has a weight corresponding for 75\% to a characteristic of $S$, and 25\% to the length of the word:
$$
\textrm{weight}_w = 75\% / |S| + 25 \% * |w| / (\sum_{w_i \in S} |w_i|)
$$
The search starts generating a random couple of prefixes and suffixes sets ($S_p$,$S_s$), i.e., for each word $w$ of $S$ an integer is selected for splitting $w$ into a prefix $p$ and a suffix $s$ such that $w=p.s$. Hence, at each iteration, the best couple $(p,s)$ is found for the selected word $w$. This process is iterated until a maximum number of iterations is reached.

In our ILS, it is not necessary to introduce noise with random walks or restarts because our process of selection of word naturally ensures diversification.

\begin{algorithm}[htb]
\caption{Iterated Local Search}\label{alg:ILS}
\renewcommand{\algorithmiccomment}[1]{\qquad\qquad // #1}
\renewcommand{\algorithmicrequire}{\textbf{Input:}}
\renewcommand{\algorithmicensure}{\textbf{Output:}}
\begin{algorithmic}[1]
    \REQUIRE set of words $S$,  
            maximum number of iterations $max\_iter$\\
            maximum of consecutive iterations allowed without improvement $max\_iter\_without\_improv$
    \ENSURE set of prefixes $S^*_p$, set of suffixes $S^*_s$
	\STATE Couple of prefixes and suffixes sets ($S_p$,$S_s$) is randomly generated
	\STATE ($S^*_p$,$S^*_s$) = ($S_p$,$S_s$) 
	\REPEAT
    	\STATE Choose a word $w$ in $S$ with a roulette wheel selection
    	\STATE ($S_p$,$S_s$) is updated by the best couple of the sub-search space corresponding only to a modification of the prefix and the suffix of word $w$ 
		\IF{$fitness(S_p,S_s) < fitness(S^*_p,S^*_s)$} 
		    \STATE ($S^*_p$,$S^*_s$) = ($S_p$,$S_s$)         
        \ENDIF
	\UNTIL{maximum number of iterations $max\_iter$ is reached or ($S^*_p$,$S^*_s$) is not improved since $max\_iter\_without\_improv$ iterations}
    \RETURN{$(S^*_p, S^*_s)$}
\end{algorithmic}
\end{algorithm}

\subsection{Genetic Algorithm Hybrid Model $HM\_GA_k$}
\label{subsec:GAHM}

We propose a classical genetic algorithm (GA) based on the search space and fitness function presented in Section~\ref{subsec:SSandEF}. A population of individuals, represented by a couple of prefixes and suffixes sets, is improved generation after generation. Each generation keeps a portion of individuals as parents and creates children by crossing the selected parents. Crossover operator used in our GA is the well-known uniform crossover. For each word, children inherit the prefix and the suffix of one of their parents randomly chosen. 
Since the population size is the same during all the search, we have a steady-state GA. A mutation process is applied over all individuals with a probability $p_{mut}$. For each word $w$, each prefix and suffix are randomly mutated by generating an integer $n$ between 0 and $|w|$ splitting $w$ into a new prefix of size $n$ and a new suffix $|w|-n$. The search stops when the maximum number of generations is reached or when no improvement is observed in the population during  $max\_gen\_without\_improv$ generations.

\begin{algorithm}[htb]
\caption{Genetic Algorithm}\label{alg:GA}
\renewcommand{\algorithmiccomment}[1]{\qquad\qquad // #1}
\renewcommand{\algorithmicrequire}{\textbf{Input:}}
\renewcommand{\algorithmicensure}{\textbf{Output:}}
\begin{algorithmic}[1]
    \REQUIRE set of words $S$, population size $s_{\cal{P}}$, mutation probability $p_{mut}$,\\
            maximum number of generations $max\_gen$,\\
            portion of population conserve in the next generation $p_{parents}$,\\
            maximum of consecutive generations allowed without improvement $max\_gen\_without\_improv$
    \ENSURE set of prefixes $S^*_p$, set of suffixes $S^*_s$
	\STATE Population $\cal{P}$ of couples of prefixes and suffixes sets ($S_p$,$S_s$) is randomly generated
	\STATE ($S^*_p$,$S^*_s$) = Argmin$_{fitness}$($\cal{P}$)
	\REPEAT
    	\STATE Select as parents set $Par$ a portion $p_{parents}$ of $\cal{P}$
    	\STATE Generate $(1-p_{parents}).s_{\cal{P}}$ children by uniform crossover over parents in a set $Children$
    	\STATE ${\cal{P}} = Par \cup Children$
    	\STATE Mutate for each individual of $\cal{P}$ the prefix/suffix for each words of $S$ with a probability $p_{mut}$
        \STATE Update the population
	    \STATE Update ($S^*_p$,$S^*_s$) if necessary         
	\UNTIL{maximum number of generations $max\_gen$ is reached or ($S^*_p$,$S^*_s$) is not improved since $max\_gen\_without\_improv$ generations}
    \RETURN{$S^*_p$ and $S^*_s$}
\end{algorithmic}
\end{algorithm}

\section{Experimental results}
\label{sec:Expe}

To test our new models, we work on the training set of the StaMinA Competition (see http://stamina.chefbe.net). We use 11 of the instances selected in \cite{DBLP:journals/fuin/JastrzabCW21}\footnote{We kept the "official" name used in \cite{DBLP:journals/fuin/JastrzabCW21}.} with a sparsity $s \in \{12.5\%, 25\%, 50\%, 100\%\}$ and an alphabet size $|\Sigma| \in \{2, 5, 10\}$. We try to generate SAT instances for NFA sizes ($k$) near to the threshold of the existence or not of an NFA.

\subsection{Experimental Protocol}
\label{subsec:protocol}
All our algorithms are implemented in Python using specific libraries such as Pysat. The experiments were carried out on a computing cluster with Intel-E5-2695 CPUs, and a limit of 10 GB of memory was fixed. Running times were limited to 10 minutes, including generation of the model and solving time. We used the Glucose~\cite{glucose} SAT solver with the default options. For stochastic methods (ILS and GA), 30 runs are realized to exploit the results statistically.

Parameters used for our hybrid models are:
\begin{center}
\begin{tabular}{|l|r||l|r|}
\hline
\multicolumn{2}{|c||}{ILS}&\multicolumn{2}{c|}{AG}\\
\hline
$max\_iter$&10 000&$s_{\cal{P}}$&100\\
\hline
$max\_iter\_without\_improv$&100&$max\_gen$&3000\\
\hline
&&$max\_gen\_without\_improv$&100\\
\hline
&&$p_{mut}$&0.05\\
\hline
&&$p_{parents}$&0.03\\
\hline
\end{tabular}
\end{center}
\subsection{Results}
\label{subsec:results}

Our experiments are reported in Table \ref{table:results}. The first column ($Instance$) corresponds to the official name of the instance, and the second one ($k$) to the number of states of the expected NFA. Then, we have in sequence the model name ($Model$), the number of SAT variables ($Var.$), the number of clauses ($Cl.$), and the instance generation time ($t_M$). The right part of the table corresponds to the solving part with the satisfiability of the generated instance ($SAT$), the decisions number ($Dec.$), and the solving time ($t_S$) with Glucose. Finally, the last column ($t_T$) corresponds to the total time (modeling time + solving time).
Results for hybrid models based on ILS ($HM\_ILS_k$) and GA ($HM\_GA_k$) correspond to average values over 30 runs. We have decided to only provide the average since the standard deviation values are very small.  

The last lines of the table correspond to the cumulative values for each column and each model. When an instance is not solved (time-out), the maximum value needed for solving the other model instances is considered. For the instance generation time ($t_M$), a credit of $600$ seconds is applied when generation did not succeed before the time-out. 

\smallskip

We can clearly confirm that the direct model is not usable in practice, and that instances cannot be generated in less than $600 s.$ The prefix model allows the fastest generation when it terminates before the time out (on these benchmarks, it did not succeed once and was thus penalize for cumulative values). It also provides instances that are solved quite fast. As expected, the instances optimized with GA are the smallest ones. However, the generation is too costly: the gain in solving time is not sufficient to compensate the long generation time. In total, in terms of solving+generation time, GA based model is close to prefix model.
As planned with its space complexity (in ${\cal{O}}(k^3)$), suffix based instances are huge and long to solve. However, we were surprised for 2 benchmarks (ww-10-40 and ww-10-50) for which the generated instances are relatively big (5 times the size of the GA optimized instances), but their solving is the fastest. We still cannot explain what made these instances easy to solve, and we are still investigating their structure.
The better balance is given with the ILS model: instances are relatively small, the generation time is fast, and the solving time as well. This is thus the best option of this work.

It is very difficult to compare our results with the results of~\cite{DBLP:journals/fuin/JastrzabCW21}. First of all, in~\cite{DBLP:journals/fuin/JastrzabCW21}, they try to minimize $k$, the number of states. Moreover, they use parallel algorithms. Finally, they do not detail the results for each instance and each $k$, except for st-2-30 and st-5-50. For the first one, with $k=9$ we are much faster. But for the second one, with $k=5$ we are slower.

\begin{table}[ht!]
\setlength{\tabcolsep}{4pt}
\centering
\caption{Comparison on 11 instances between the models $DM_k$, $PM_k$, $SM_k$, $HM\_ILS_k$, and $HM\_GA_k$.}
\label{table:results}
\begin{scriptsize}
\renewcommand{\arraystretch}{1.1}
\begin{tabular}{|l|c|l|r|r|r||c|r|r||r|}
\hline
Instance&k&Model&Var.&Cl.&$t_M$&SAT&Dec.&$t_S$&$t_T$\\
\hline
	&		&	$DM_k$	&	190 564	&	1 817 771	&	13.46	&	True	&	973 213	&	88.62	&	102.09	\\ \cline{3-10}
	&		&	$PM_k$	&	1 276	&	4 250	&	1.27	&	True	&	3 471	&	0.10	&	\textbf{1.37}	\\ \cline{3-10}
st-2-10	&	4	&	$SM_k$	&	5 196	&	17 578	&	1.30	&	True	&	4 332	&	0.21	&	1.51	\\ \cline{3-10}
	&		&	$HM\_ILS_k$	&	1 188	&	4 179	&	4.14	&	True	&	2 503	&	\textbf{0.05}	&	4.18	\\ \cline{3-10}
	&		&	$HM\_GA_k$	&	\textbf{1 107}	&	3 884	&	14.45	&	True	&	2 368	&	\textbf{0.05}	&	14.50	\\ \hline
	&		&	$DM_k$	&	-	&	-	&	-	&	-	&	-	&	-	&	-	\\ \cline{3-10}
	&		&	$PM_k$	&	4 860	&	17 150	&	1,34	&	False	&	1 625 706	&	241,24	&	242,59	\\ \cline{3-10}
st-2-20	&	6	&	$SM_k$	&	-	&	-	&	-	&	-	&	-	&	-	&	-	\\ \cline{3-10}
	&		&	$HM\_ILS_k$	&	5 688	&	21 073	&	5,39	&	False	&	662 354	&	98,35	&	\textbf{103,74}	\\ \cline{3-10}
	&		&	$HM\_GA_k$	&	\textbf{4 735}	&	17 611	&	34,65	&	False	&	708 356	&	\textbf{94,95}	&	129,61	\\ \hline
	&		&	$DM_k$	&	-	&	-	&	-	&	-	&	-	&	-	&	-	\\ \cline{3-10}
	&		&	$PM_k$	&	-	&	-	&	-	&	-	&	-	&	-	&	-	\\ \cline{3-10}
st-2-30	&	9	&	$SM_k$	&	-	&	-	&	-	&	-	&	-	&	-	&	-	\\ \cline{3-10}
	&		&	$HM\_ILS_k$	&	20 637	&	78 852	&	7.55	&	True	&	1 998 574	&	\textbf{228.53}	&	\textbf{236.07}	\\ \cline{3-10}
	&		&	$HM\_GA_k$	&	\textbf{16 335}	&	62 832	&	66.94	&	True	&	4 079 686	&	527.44	&	594.38	\\ \hline
	&		&	$DM_k$	&	-	&	-	&	-	&	-	&	-	&	-	&	-	\\ \cline{3-10}
	&		&	$PM_k$	&	4 024	&	13 464	&	1.49	&	True	&	2 641	&	\textbf{0.08}	&	\textbf{1.57}	\\ \cline{3-10}
st-5-20	&	4	&	$SM_k$	&	14 964	&	50 660	&	1.68	&	True	&	23 540	&	3.23	&	4.91	\\ \cline{3-10}
	&		&	$HM\_ILS_k$	&	3 608	&	12 514	&	7.83	&	True	&	14 584	&	0.72	&	8.56	\\ \cline{3-10}
	&		&	$HM\_GA_k$	&	\textbf{3 522}	&	12 180	&	47.84	&	True	&	18 344	&	0.94	&	48.78	\\ \hline
	&		&	$DM_k$	&	-	&	-	&	-	&	-	&	-	&	-	&	-	\\ \cline{3-10}
	&		&	$PM_k$	&	5 364	&	18 054	&	1.43	&	True	&	177 711	&	21.57	&	\textbf{23.00}	\\ \cline{3-10}
st-5-30	&	4	&	$SM_k$	&	21 084	&	71 502	&	1.87	&	True	&	362 318	&	128.02	&	129.89	\\ \cline{3-10}
	&		&	$HM\_ILS_k$	&	4 837	&	16 955	&	9.90	&	True	&	156 631	&	\textbf{19.90}	&	29.81	\\ \cline{3-10}
	&		&	$HM\_GA_k$	&	\textbf{4 705}	&	16 478	&	119.42	&	True	&	171 062	&	21.67	&	141.09	\\ \hline
	&		&	$DM_k$	&	-	&	-	&	-	&	-	&	-	&	-	&	-	\\ \cline{3-10}
	&		&	$PM_k$	&	6 284	&	21 216	&	1.52	&	False	&	7 110	&	0.55	&	\textbf{2.07}	\\ \cline{3-10}
st-5-40	&	4	&	$SM_k$	&	23 604	&	80 104	&	1.55	&	False	&	15 708	&	1.74	&	3.29	\\ \cline{3-10}
	&		&	$HM\_ILS_k$	&	5 745	&	20 290	&	10.29	&	False	&	6 206	&	\textbf{0.34}	&	10.62	\\ \cline{3-10}
	&		&	$HM\_GA_k$	&	\textbf{5 548}	&	19 517	&	150.50	&	False	&	6 204	&	0.35	&	150.85	\\ \hline
	&		&	$DM_k$	&	-	&	-	&	-	&	-	&	-	&	-	&	-	\\ \cline{3-10}
	&		&	$PM_k$	&	11 150	&	38 745	&	1.59	&	False	&	1 943 735	&	562.80	&	564.39	\\ \cline{3-10}
st-5-50	&	5	&	$SM_k$	&	-	&	-	&	-	&	-	&	-	&	-	&	-	\\ \cline{3-10}
	&		&	$HM\_ILS_k$	&	11 085	&	40 258	&	10.80	&	False	&	911 280	&	\textbf{238.10}	&	\textbf{248.90}	\\ \cline{3-10}
	&		&	$HM\_GA_k$	&	\textbf{10 040}	&	36 350	&	279.87	&	False	&	1 093 093	&	287.46	&	567.33	\\ \hline
	&		&	$DM_k$	&	-	&	-	&	-	&	-	&	-	&	-	&	-	\\ \cline{3-10}
	&		&	$PM_k$	&	14 200	&	49 455	&	1.52	&	False	&	1 245 538	&	383.37	&	384.89	\\ \cline{3-10}
st-5-60	&	5	&	$SM_k$	&	-	&	-	&	-	&	-	&	-	&	-	&	-	\\ \cline{3-10}
	&		&	$HM\_ILS_k$	&	13 920	&	50 568	&	13.47	&	False	&	800 920	&	\textbf{231.82}	&	\textbf{245.29}	\\ \cline{3-10}
	&		&	$HM\_GA_k$	&	\textbf{13 180}	&	47 755	&	313.30	&	False	&	950 601	&	270.97	&	584.26	\\ \hline
	&		&	$DM_k$	&	15 012	&	112 039	&	2.07	&	True	&	69 219	&	1.52	&	3.59	\\ \cline{3-10}
	&		&	$PM_k$	&	3 624	&	11 900	&	1.38	&	True	&	977	&	0.03	&	1.41	\\ \cline{3-10}
ww-10-40	&	4	&	$SM_k$	&	13 844	&	46 648	&	1.25	&	True	&	4 173	&	\textbf{0.02}	&	\textbf{1.28}	\\ \cline{3-10}
	&		&	$HM\_ILS_k$	&	2 896	&	10 342	&	5.94	&	True	&	3 897	&	0.06	&	6.00	\\ \cline{3-10}
	&		&	$HM\_GA_k$	&	\textbf{2 761}	&	9 839	&	75.57	&	True	&	2 842	&	0.04	&	75.60	\\ \hline
	&		&	$DM_k$	&	80 548	&	694 641	&	5.61	&	True	&	483 153	&	103.52	&	109.14	\\ \cline{3-10}
	&		&	$PM_k$	&	5 364	&	17 850	&	1.28	&	True	&	167 390	&	20.29	&	21.57	\\ \cline{3-10}
ww-10-50	&	4	&	$SM_k$	&	20 844	&	70 482	&	1.49	&	True	&	74 482	&	11.58	&	\textbf{13.07}	\\ \cline{3-10}
	&		&	$HM\_ILS_k$	&	4 633	&	16 514	&	7.71	&	True	&	73 534	&	5.46	&	13.17	\\ \cline{3-10}
	&		&	$HM\_GA_k$	&	\textbf{4 517}	&	15 940	&	123.21	&	True	&	52 894	&	\textbf{3.38}	&	126.59	\\ 
	\hline 
	\multicolumn{10}{c}{~}\\ 
	\hline

\multicolumn{2}{|l|}{}			&	$DM_k$	&	397 451	&	3 014 842	&	4 221,15	&	-	&	10 821 816	&	2 041,50	&	6 262,65	\\ \cline{3-10}
\multicolumn{2}{|l|}{}			&	$PM_k$	&	76 783	&	270 936	&	612,82	&	-	&	9 253 965	&	1 757,47	&	2 370,29	\\ \cline{3-10}
\multicolumn{2}{|l|}{Cumulative values}			&	$SM_k$	&	151 211	&	525 099	&	2 409,15	&	-	&	9 379 218	&	1 879,65	&	4 268,80	\\ \cline{3-10}
\multicolumn{2}{|l|}{}			&	$HM\_ILS_k$	&	74 237	&	271 543	&	\textbf{83,03}	&	-	&	4 630 483	&	\textbf{823,33}	&	\textbf{906,36}	\\ \cline{3-10}
\multicolumn{2}{|l|}{}			&	$HM\_GA_k$	&	\textbf{66 450}	&	242 386	&	1 225,75	&	-	&	7 085 451	&	1 207,23	&	2 432,98	\\ \hline

\hline
\end{tabular}
\end{scriptsize}
\end{table}

\section{Conclusion}
\label{sec:Conclusion}
In this paper, we have proposed to use some metaheuristics algorithms, namely ILS and GA, to improve the size of SAT models for the NFA inferring problem. Our hybrid model, optimized with GA gives, on average, the smallest SAT instances. Solving these instances is also faster than with the direct or prefix models. However, generation of the optimized instances with GA is really too long and is not balanced out with the gain in solving time; it is at the level of the prefix model w.r.t. total CPU time. The ILS model generates optimized instances a bit larger than with GA and a bit smaller than with prefixes. Moreover, the solving time is the best of our experiments, and the generation time added to the solving time makes of the $HM\_ILS_k$ our better model. 

In the future, we plan to speed up GA to make it more competitive. We also plan to consider more complex fitness functions, not only based on the number of SAT variables but also on the length of clauses. We also plan a model portfolio approach for larger samples.

\bibliographystyle{splncs}
\bibliography{biblio}

%

\end{document}